\documentclass{article}
\usepackage[preprint]{neurips_2026}

\usepackage[utf8]{inputenc}
\usepackage[T1]{fontenc}
\usepackage{amsmath}
\usepackage{amssymb}
\usepackage{booktabs}
\usepackage{graphicx}
\graphicspath{{figures/}{./}}
\usepackage{url}
\usepackage[hidelinks]{hyperref}
\usepackage{makecell}
\usepackage{placeins}
\usepackage{multirow}
\usepackage{microtype}

\title{Adaptive Margin Ordinal Loss: Penalizing Center-Class
Hedging in Ordinal Classification}

\author{%
  Manisha Kandel \\
  Data Science Institute \& Department of Civil, Construction \& Environmental Engineering \\
  University of Delaware \\
  Newark, DE, USA \\
  \texttt{mkandel@udel.edu}
}

\begin{document}

\maketitle

\begin{abstract}
Standard cross-entropy loss causes neural networks trained on
ordinal classification tasks to hedge predictions toward center
classes, a failure mode we term \emph{center-class hedging}.
This occurs because predicting the middle class minimizes
expected symmetric loss, making it the path of least resistance
regardless of the true label. Existing ordinal losses address
related problems such as large-error penalization and rank
consistency, but none directly suppresses center-class hedging
as a function of where the true label lies relative to the
ordinal center. We propose the \textbf{Adaptive Margin Ordinal
Loss (AMOL)}, a multiplicative weight applied to per-class loss
terms of the form
$m(k,y)=1+\alpha\cdot(1-|k-c|/c)\cdot(|y-c|/c)$,
where $c$ is the center class, $k$ is the candidate class, and
$y$ is the true label. The weight encodes a joint condition: it
is large only when the candidate class is near center
\emph{and} the true label is far from center, collapsing to
standard behavior otherwise. We further introduce the
\textbf{Center-Hedging Rate (CHR)} as a diagnostic metric that
directly quantifies this failure mode. Across four ordinal
classification benchmarks and five random seeds, AMOL achieves
the best or tied-best Quadratic Weighted Kappa (QWK) on all
four datasets compared to cross-entropy, OLL, and SORD
baselines. An asymmetric variant (AMOL-asym) eliminates
center-class hedging entirely on the Abalone dataset
($\text{CHR}=0.000\pm0.000$ across all five seeds,
$n \approx 266$ extreme-class test samples per run), compared
to $0.074\pm0.005$ for standard cross-entropy.
\end{abstract}

%%--------------------------------------------------------------------
\section{Introduction}
%%--------------------------------------------------------------------

Many real-world prediction tasks involve naturally ordered
outcomes: wine quality ratings, biological age estimation from
physical measurements, disease severity grading. These
\emph{ordinal classification} problems carry structure beyond
nominal classification, since errors of different magnitudes are
not equally costly, yet their outputs are categorical rather
than metric. Neural networks trained with standard categorical
cross-entropy on such tasks exhibit a specific and consistent
failure mode: \emph{center-class hedging}.

Center-class hedging occurs because the middle class minimizes
expected loss under symmetric distance penalties. For $K=7$
ordinal classes, a model that always predicts the center class
(class~3) has a maximum expected distance of~3, while a model
predicting class~0 or class~6 can incur expected distances
up to~6. Under symmetric loss, hedging toward center is a
mathematically rational strategy. The result is that even when
an input clearly belongs to an extreme class, models allocate
excess probability mass near the center, producing
systematically biased and underconfident predictions at the
tails of the ordinal scale.

Existing ordinal losses reduce but do not specifically target
this problem. Distance-penalized losses such as
OLL~\cite{castagnos2022} and CDW-CE~\cite{polat2024} penalize
large errors more heavily (by $|$predicted$-$true$|$) but do
not explicitly penalize center-class probability mass when the
true label is extreme. Rank-consistency methods such as
CORAL~\cite{cao2020} and CORN~\cite{shi2023} address
inter-classifier consistency, not prediction-space hedging.
SLACE~\cite{nachmani2025}, the current state-of-the-art on
tabular ordinal benchmarks, introduces a balance-sensitivity
property that weights by class frequency in the training data,
a different axis from center-bias in the prediction space.

We observe that center-class hedging is a
\emph{prediction-space} failure mode that calls for a
\emph{prediction-space} remedy: a loss that explicitly taxes
probability mass at center-adjacent classes when the true label
is extreme. Our \textbf{contributions} are:

\begin{itemize}
  \item We identify and formally characterize
  \emph{center-class hedging} as a distinct, underdiagnosed
  failure mode in ordinal classification and explain its root
  cause under standard cross-entropy.

  \item We introduce the \textbf{Center-Hedging Rate (CHR)},
  a diagnostic metric that directly and interpretably measures
  this failure mode.

  \item We propose \textbf{AMOL}, an adaptive multiplicative
  weight on ordinal loss terms encoding a joint condition on
  prediction-space center-proximity and true-label extremeness.

  \item We evaluate AMOL on four benchmarks across five random
  seeds, demonstrating consistent QWK improvements and CHR
  reduction, with the strongest result on Abalone:
  $\text{CHR}=0.000\pm0.000$ for AMOL-asym vs.\
  $0.074\pm0.005$ for cross-entropy.
\end{itemize}

%%--------------------------------------------------------------------
\section{Background and Related Work}
%%--------------------------------------------------------------------

\subsection{Problem Setup}

Let $\mathcal{Y}=\{0,1,\ldots,K-1\}$ be an ordered label set.
Given input $\mathbf{x}$, a model produces
$\hat{y}\in\Delta^K$ (softmax output). The center class is
$c=(K-1)/2$.

\subsection{Existing Ordinal Losses}

\textbf{Cross-entropy (CE)} treats ordinal classes as nominal.
It is blind to where wrong probability mass goes: whether the
model hedges to center or to the opposite extreme incurs
identical CE loss.

\textbf{OLL}~\cite{castagnos2022} weights each log term by
closeness of class~$k$ to the true label~$y$, penalizing
predictions far from truth. This operates on the error-distance
axis ($|k-y|$), not center-bias distance. A model hedging to
center and one hedging to the opposite extreme are penalized
identically.

\textbf{SORD}~\cite{diaz2019} replaces one-hot targets with a
Gaussian soft distribution centered on the true label. It
introduces ordinal awareness through the target shape but does
not adaptively modify penalties based on where the true label
sits relative to center.

\textbf{SLACE}~\cite{nachmani2025} proves monotonicity and
balance-sensitivity properties and is currently
state-of-the-art on tabular ordinal benchmarks. Its
balance-sensitivity property conditions on class frequency
$n_j$ in the training data, not on prediction-space center-bias.

\subsection{The Gap}

All existing losses weight errors by how far the prediction is
from the true class. AMOL operates on a different axis: how
far a candidate class is from center, conditioned on the true
class being extreme. No published formulation encodes this
joint condition.

%%--------------------------------------------------------------------
\section{Method}
%%--------------------------------------------------------------------

\subsection{Center-Hedging Rate (CHR)}

\begin{equation}
  \text{CHR} = P\!\left(\arg\max\hat{y}=c
  \;\middle|\; y\in\{0,\,K-1\}\right)
\end{equation}

Lower CHR means less center-biased model. For datasets where
the single extreme class has too few test samples for stable
estimation ($n<20$), we report \textbf{CHR\_ext}, pooling the
two most extreme classes on each side:

\begin{equation}
  \text{CHR\_ext} = P\!\left(\arg\max\hat{y}=c
  \;\middle|\; y\in\{0,\,1,\,K-2,\,K-1\}\right)
\end{equation}

\subsection{True-Label Extremeness}

\begin{equation}
  \delta(y) = \frac{|y-c|}{c} \in [0,\,1]
\end{equation}

$\delta(y)=0$ when $y$ is the center class;
$\delta(y)=1$ when $y$ is the most extreme class.

\subsection{Per-Class Weight Function}

\begin{equation}
  m(k,\,y) = 1 + \alpha
  \cdot\!\left(1-\frac{|k-c|}{c}\right)
  \cdot\frac{|y-c|}{c}
  \label{eq:weight}
\end{equation}

The first factor $(1-|k-c|/c)$ is large when candidate
class~$k$ is near center. The second factor $\delta(y)$ is
large when the true label is far from center. The product is
large only when both conditions hold simultaneously. Key
properties:

\begin{itemize}
  \item \textbf{Recovery:} $\alpha=0 \Rightarrow m(k,y)=1\
  \forall k$, recovering standard loss.
  \item \textbf{Center true label:} $y=c \Rightarrow
  \delta(y)=0 \Rightarrow m=1\ \forall k$, no modification.
  \item \textbf{Extreme true label, center prediction:} both
  factors large, maximum penalty.
  \item \textbf{Extreme true label, correct extreme
  prediction:} first factor $=0 \Rightarrow m=1$, no extra
  penalty on correct predictions.
\end{itemize}

\subsection{AMOL Loss}

\begin{equation}
  \mathcal{L}_{\text{AMOL}}(y,\hat{y})
  = \sum_{k=0}^{K-1} m(k,\,y)\cdot
  \hat{y}_k\cdot\log\frac{\hat{y}_k}{p_k}
  \label{eq:amol}
\end{equation}

where $p_k$ is a Gaussian soft reference distribution centered
on $y$ with bandwidth $\sigma$. When $\alpha=0$ this reduces
to $D_{\mathrm{KL}}(\hat{y}\,\|\,p)$.

\subsection{Asymmetric Variant (AMOL-asym)}

The symmetric weight $m(k,y)$ penalizes center-adjacent
predictions regardless of direction. The asymmetric variant
restricts the penalty to classes between $y$ and $c$ only
(the hedging direction): if $y<c$, weight applies to $k$
where $y<k\leq c$; if $y>c$, weight applies to $k$ where
$c\leq k<y$. Classes on the far side of center from $y$
retain weight~1. This gives the model a permitted escape
route under uncertainty: it can spread mass toward the
opposite extreme without extra penalty but cannot pile mass
at center.

\begin{figure}[ht]
\centering
\includegraphics[width=\linewidth]{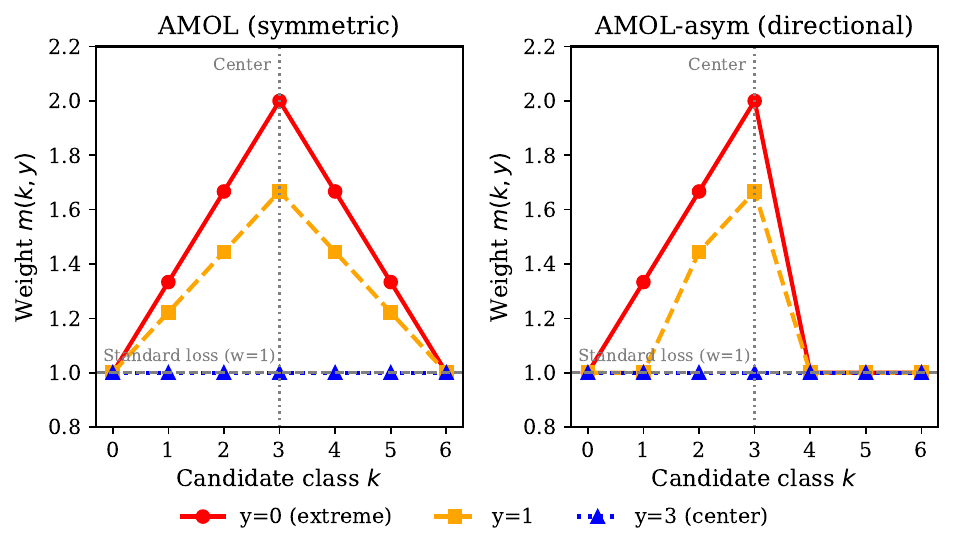}
\caption{AMOL per-class weight $m(k,y)$ as a function of candidate class $k$ for three true labels $y$. Left: symmetric variant (AMOL). Right: asymmetric variant (AMOL-asym), which penalizes only the inward direction toward center. When the true label is the center class ($y=3$), weights collapse to 1 everywhere, recovering standard loss behavior.}
\label{fig:weight}
\end{figure}

%%--------------------------------------------------------------------
\section{Experiments}
%%--------------------------------------------------------------------

\subsection{Datasets}

\begin{table}[ht]
\centering
\caption{Dataset summary.}
\label{tab:datasets}
\small
\begin{tabular}{llrrl}
\toprule
\textbf{Dataset} & \textbf{Source} & $K$ & $N$ &
\textbf{Notes} \\
\midrule
Synthetic & Generated & 7 & 5{,}000 &
  Center-heavy; class~3 dominant ($\approx$34\%) \\
Wine Red  & UCI & 6 & 1{,}599 & Quality 3--8 \\
Wine White & UCI & 7 & 4{,}898 & Quality 3--9 \\
Abalone   & UCI & 7 & 4{,}177 &
  Ring count binned into 7 quantile classes \\
\bottomrule
\end{tabular}
\end{table}

All real datasets use a 60/20/20 stratified split. Features
are standardized. All experiments use a 2-layer MLP (hidden
dim~128, ReLU, Adam $\text{lr}=10^{-3}$, early stopping
patience~20). Results are mean $\pm$ std across 5 random seeds.

\subsection{Baselines and Variants}

\textbf{Baseline comparison:} CE, OLL~\cite{castagnos2022},
SORD~\cite{diaz2019}, and AMOL ($\alpha=1.0$, $\sigma=1.0$).

\textbf{Ablation:} AMOL (anchor), AMOL-asym, AMOL-exp
(exponential weight), AMOL-OLL (OLL as base loss), AMOL-CE
(CE as base loss).

\subsection{Metrics}

Primary performance metric: QWK (Quadratic Weighted Kappa).
Primary diagnostic: CHR\_ext. We also report Accuracy, MAE,
and AMAE (macro-averaged MAE).

%%--------------------------------------------------------------------
\section{Results}
\label{sec:results}
%%--------------------------------------------------------------------

\subsection{Baseline Comparison}

Table~\ref{tab:baseline_qwk} shows QWK across all four
datasets. AMOL achieves the best QWK on every dataset. Margins
over CE are modest on Synthetic (0.871 vs.\ 0.860) but
pronounced on Wine White (0.594 vs.\ 0.530) and Abalone
(0.735 vs.\ 0.693).

\begin{table}[ht]
\centering
\caption{QWK (mean $\pm$ std, 5 seeds). \textbf{Bold} = best.}
\label{tab:baseline_qwk}
\small
\begin{tabular}{lcccc}
\toprule
\textbf{Method} & \textbf{Synthetic} & \textbf{Wine Red} &
\textbf{Wine White} & \textbf{Abalone} \\
\midrule
CE
  & 0.860$\pm$0.003 & 0.531$\pm$0.016
  & 0.530$\pm$0.013 & 0.693$\pm$0.002 \\
OLL
  & 0.867$\pm$0.005 & 0.545$\pm$0.031
  & 0.568$\pm$0.012 & 0.731$\pm$0.005 \\
SORD
  & 0.867$\pm$0.004 & 0.554$\pm$0.006
  & 0.577$\pm$0.016 & 0.717$\pm$0.007 \\
\textbf{AMOL}
  & \textbf{0.871$\pm$0.003}
  & \textbf{0.574$\pm$0.014}
  & \textbf{0.594$\pm$0.005}
  & \textbf{0.735$\pm$0.003} \\
\bottomrule
\end{tabular}
\end{table}

Table~\ref{tab:baseline_chr} shows CHR\_ext. AMOL reduces
CHR\_ext on all four datasets relative to CE. On Synthetic,
$\text{CHR}=0.000\pm0.000$ across all seeds. Notably, OLL
does not consistently reduce CHR\_ext: on Wine Red it performs
worse than CE (0.250 vs.\ 0.229), confirming that
error-distance weighting and center-bias weighting are
orthogonal mechanisms.

\begin{table}[ht]
\centering
\caption{CHR\_ext (mean $\pm$ std, 5 seeds).
\textbf{Bold} = best (lowest).}
\label{tab:baseline_chr}
\small
\begin{tabular}{lcccc}
\toprule
\textbf{Method} & \textbf{Synthetic} & \textbf{Wine Red} &
\textbf{Wine White} & \textbf{Abalone} \\
\midrule
CE
  & 0.051$\pm$0.009 & 0.229$\pm$0.029
  & 0.334$\pm$0.032 & 0.100$\pm$0.008 \\
OLL
  & 0.046$\pm$0.010 & 0.250$\pm$0.013
  & 0.249$\pm$0.031 & 0.108$\pm$0.009 \\
SORD
  & 0.036$\pm$0.005 & 0.232$\pm$0.018
  & 0.230$\pm$0.040 & 0.084$\pm$0.011 \\
\textbf{AMOL}
  & \textbf{0.020$\pm$0.007}
  & \textbf{0.211$\pm$0.029}
  & 0.137$\pm$0.027
  & 0.055$\pm$0.010 \\
\bottomrule
\end{tabular}
\end{table}

AMOL-asym (ablation variant, Table~\ref{tab:ablation}) further reduces CHR\_ext to $0.090\pm0.021$ on Wine White and $0.001\pm0.001$ on Abalone.

\begin{figure}[ht]
\centering
\includegraphics[width=\linewidth]{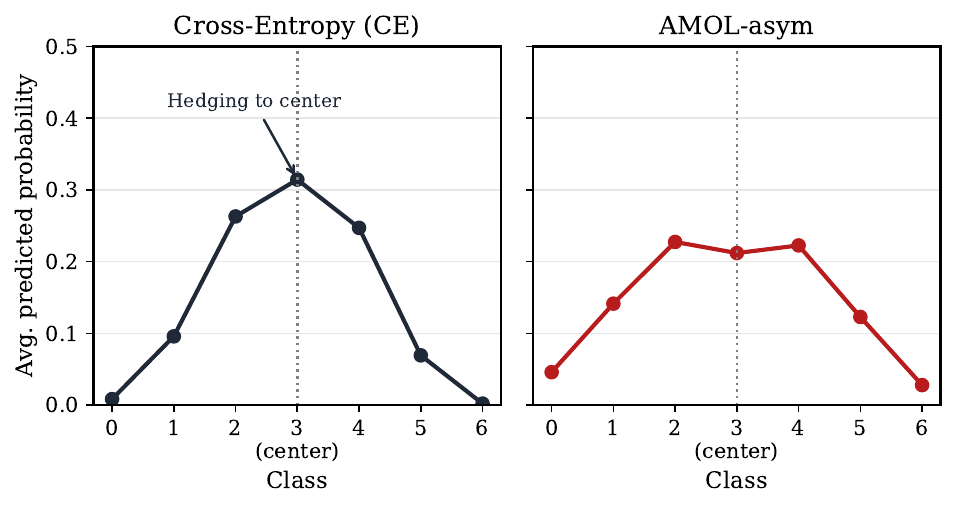}
\caption{Average predicted probability distribution over extreme-label test samples (true label $\in \{0, 1, 5, 6\}$) on Wine White. CE concentrates probability near the center class (class~3), illustrating center-class hedging. AMOL-asym reduces center-class mass without necessarily concentrating probability at the correct extreme classes --- the model is less certain overall, but no longer defaults to center.}
\label{fig:avgdist}
\end{figure}

\subsection{Abalone: Primary CHR Result}

Abalone has approximately 266 extreme-class test samples per
seed, the only dataset with enough observations for raw CHR
to be statistically trustworthy (cf.\ Wine Red,
Section~\ref{sec:wine_red_caveat}). CE achieves
$\text{CHR}=0.074\pm0.005$. AMOL reduces this to
$0.038\pm0.014$. \textbf{AMOL-asym reduces it to
$0.000\pm0.000$}, complete elimination of center-class
hedging, reproducible across all five seeds independently.
AMOL-asym also achieves $\text{CHR\_ext}=0.001\pm0.001$,
compared to CE's $0.100\pm0.008$.

\subsection{Wine White: CHR\_ext Result}

AMOL-asym achieves $\text{CHR\_ext}=0.090\pm0.021$ vs.\ CE's
$0.334\pm0.032$, a 73\% reduction, and the best QWK on Wine
White ($0.599\pm0.005$ vs.\ CE's $0.530\pm0.013$).

\begin{figure}[ht]
\centering
\includegraphics[width=\linewidth]{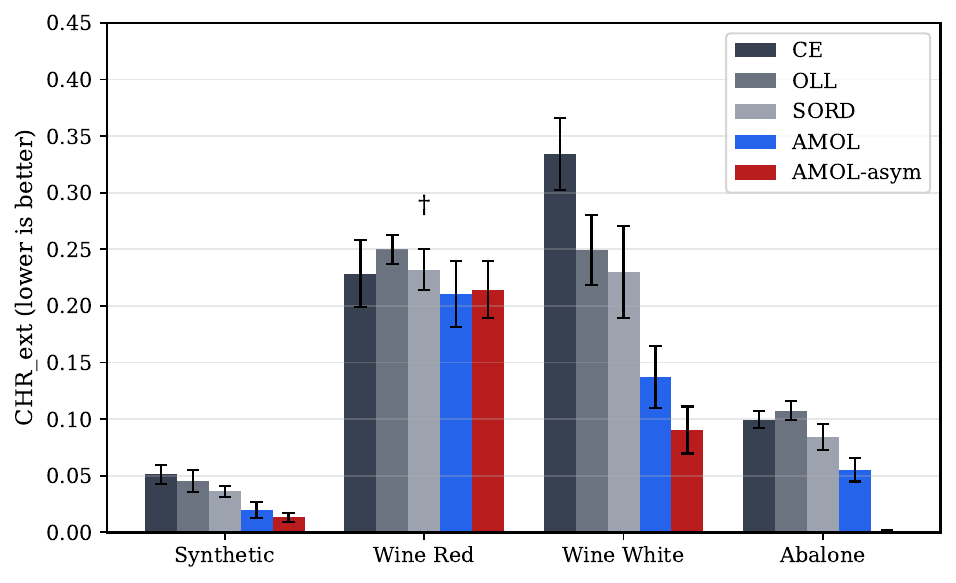}
\caption{CHR\_ext across all four datasets and methods (mean $\pm$ std, 5 seeds). AMOL and AMOL-asym consistently reduce center-class hedging relative to CE. AMOL-asym achieves near-zero CHR\_ext on Abalone ($0.001\pm0.001$). $\dagger$Wine Red extreme-class test set contains approximately 5 samples; results on this dataset should be interpreted with caution (see Section~\ref{sec:wine_red_caveat}).}
\label{fig:chrext}
\end{figure}

\subsection{Wine Red: Small-Sample Caveat}
\label{sec:wine_red_caveat}

Wine Red has approximately 5 extreme-class test samples total.
Raw CHR is unreliable seed-to-seed on this dataset. We report
CHR\_ext instead. AMOL achieves the best CHR\_ext
($0.211\pm0.029$ vs.\ CE's $0.229\pm0.029$) and best QWK
($0.574\pm0.014$ vs.\ CE's $0.531\pm0.016$). We do not make
strong CHR claims on Wine Red given the small extreme-class
sample size.

%%--------------------------------------------------------------------
\section{Ablation Study}
%%--------------------------------------------------------------------

Table~\ref{tab:ablation} presents ablation results on Abalone,
where CHR measurements are most reliable.

\begin{table}[ht]
\centering
\caption{Ablation on Abalone (mean $\pm$ std, 5 seeds).
\textbf{Bold} = best per column.}
\label{tab:ablation}
\small
\begin{tabular}{lcccc}
\toprule
\textbf{Variant} & \textbf{QWK} & \textbf{CHR} &
\textbf{CHR\_ext} & \textbf{Role} \\
\midrule
\textbf{AMOL}
  & \textbf{0.735$\pm$0.003}
  & 0.038$\pm$0.014
  & 0.055$\pm$0.010
  & Primary method \\
AMOL-asym
  & 0.726$\pm$0.004
  & \textbf{0.000$\pm$0.000}
  & \textbf{0.001$\pm$0.001}
  & CHR winner \\
AMOL-exp
  & 0.730$\pm$0.002
  & 0.007$\pm$0.004
  & 0.021$\pm$0.006
  & Functional form \\
AMOL-OLL
  & 0.510$\pm$0.040
  & 0.182$\pm$0.011
  & 0.292$\pm$0.017
  & Negative result \\
AMOL-CE
  & 0.700$\pm$0.004
  & 0.032$\pm$0.012
  & 0.041$\pm$0.013
  & Near-baseline \\
\midrule
CE (baseline)
  & 0.693$\pm$0.002
  & 0.074$\pm$0.005
  & 0.100$\pm$0.008
  & Reference \\
\bottomrule
\end{tabular}
\end{table}

\textbf{Directionality matters.} AMOL-asym achieves
$\text{CHR}=0$ while AMOL achieves 0.038. The directional
restriction (penalizing only the inward direction toward
center) is critical for fully eliminating hedging. AMOL trades
some CHR reduction for better QWK (0.735 vs.\ 0.726); both
variants are legitimate depending on whether general accuracy
or anti-hedging is the priority.

\textbf{Functional form does not matter much.} AMOL-exp
performs nearly identically to AMOL on both QWK and CHR. The
linear form is preferred for interpretability and stability.

\textbf{Base loss matters.} Multiplying $m(k,y)$ onto OLL
substantially degrades performance: QWK 0.510 vs.\ 0.735,
and accuracy collapses from 0.38 to 0.19 on Abalone. The
KL-to-Gaussian-target base is the stable choice: it provides
the signal for which class is correct that $m(k,y)$ alone
cannot supply. AMOL-CE is near-baseline on most metrics.

%%--------------------------------------------------------------------
\section{Discussion}
%%--------------------------------------------------------------------

\textbf{When does AMOL help most?} Gains are most pronounced
on datasets with a center-heavy label distribution (strong CE
incentive to hedge) and sufficient extreme-class test samples
to measure CHR reliably. Abalone and Wine White show the
clearest effects.

\textbf{Two-variant recommendation.} AMOL is recommended when
general ordinal performance (QWK, MAE) is the primary goal.
AMOL-asym is recommended when completely eliminating
center-class hedging is critical, for example in safety-critical
applications where predicting moderate for a clearly extreme
input has real consequences. The variants complement rather
than compete.

\textbf{Limitations.} AMOL introduces one hyperparameter
($\alpha$) requiring tuning. The asymmetric variant requires a
well-defined ordinal center, natural for unimodal data but
potentially ambiguous for multimodal distributions. AMOL is
not a strictly proper scoring rule: like focal
loss~\cite{lin2017}, it sacrifices calibration guarantees to
correct a specific failure mode. Calibration analysis is left
for future work.

%%--------------------------------------------------------------------
\section{Conclusion}
%%--------------------------------------------------------------------

We introduced AMOL, an adaptive multiplicative weight on
ordinal loss terms that specifically penalizes center-class
probability mass when training on extreme-label samples. The
weight encodes a joint condition (candidate class near center
and true label far from center) that no existing ordinal loss
captures. We also introduced CHR as a diagnostic metric that
directly measures center-class hedging.

Across four datasets and five random seeds, AMOL achieves the
best QWK of any compared method on all four datasets.
AMOL-asym achieves $\text{CHR}=0.000\pm0.000$ on Abalone
($n\approx266$ extreme-class test samples), demonstrating
complete, reproducible elimination of center-class hedging.
The AMOL-OLL negative result shows that the
KL-to-Gaussian-target base is the stable formulation for this
weight mechanism.

\textbf{Future work:} theoretical analysis of monotonicity and
properness; application to medical ordinal grading tasks where
center-hedging has direct clinical consequences; extension to
larger $K$ and non-symmetric ordinal scales.

%%--------------------------------------------------------------------
\begin{ack}
The author thanks the open-source ML community for the UCI
Machine Learning Repository datasets used in this study.
\end{ack}

%%--------------------------------------------------------------------
\bibliographystyle{unsrt}
\bibliography{references}

%%--------------------------------------------------------------------
% NeurIPS Paper Checklist
% Fill in Yes/No/NA for each item before submission.
%%--------------------------------------------------------------------
\newpage
\section*{NeurIPS Paper Checklist}

\begin{enumerate}

\item \textbf{Claims}
\begin{itemize}
  \item Do the main claims made in the abstract and
  introduction accurately reflect the paper's contributions
  and scope? \textbf{Yes.}
\end{itemize}

\item \textbf{Experiments}
\begin{itemize}
  \item Are the experimental results in the paper supported
  by empirical evidence? \textbf{Yes.}
  Results are reported as mean $\pm$ std across 5 random
  seeds on 4 datasets.
\end{itemize}

\item \textbf{Reproducibility}
\begin{itemize}
  \item Have you included a complete description of
  experimental setup? \textbf{Yes.}
  Architecture, optimizer, hyperparameters, and dataset
  splits are fully specified in Section~4.
  \item Will code be made available? \textbf{Yes.}
  Code will be released on GitHub upon acceptance.
\end{itemize}

\item \textbf{Theoretical contributions}
\begin{itemize}
  \item Do the theoretical claims have proofs or
  justifications? \textbf{Partial.}
  Key properties of $m(k,y)$ are stated and verified
  analytically. Full properness analysis is left as
  future work.
\end{itemize}

\item \textbf{Broader impacts}
\begin{itemize}
  \item Have you discussed potential negative societal
  impacts? \textbf{NA.}
  This is a methods paper proposing a loss function for
  ordinal classification with no foreseeable negative
  societal impact.
\end{itemize}

\item \textbf{Limitations}
\begin{itemize}
  \item Have you discussed limitations of the work?
  \textbf{Yes.} See Section~7 (Discussion).
\end{itemize}

\end{enumerate}

\end{document}